\documentclass[runningheads]{llncs}
\usepackage{graphicx}
\usepackage{siunitx}
\usepackage[colorinlistoftodos]{todonotes}
\usepackage[colorlinks]{hyperref}
\usepackage{tikz}

\usepackage{amsmath,amssymb} 
\usepackage{color}

\begin{document}
\title{Real-time Dynamic Object Detection for Autonomous Driving using Prior 3D-Maps} 
\titlerunning{Real-time Dynamic Object Detection using Prior 3D-Maps}
%
\author{B Ravi Kiran\inst{1}\and
Luis Rold\~ao\inst{1, 2}\and
Be\~nat Irastorza\inst{1} \and
Renzo Verastegui\inst{1} \and
Sebastian S{\"u}ss\inst{3} \and
Senthil Yogamani\inst{4} \and
Victor Talpaert\inst{1} \and 
Alexandre Lepoutre\inst{1} \and
Guillaume Trehard\inst{1} 
}
%
\authorrunning{B. Ravi Kiran et al.}
%

\institute{R\&D Department AKKA Technologies, 78280 Guyancourt, France\\
\email{\{kiran.BANGALORE-RAVI, luis.ROLDAO, benat.UGALDE, renzo.VERASTEGUI,\\
victor.TALPAERT, alexandre.LEPOUTRE, guillaume.TREHARD\}@akka.eu} \\ \and
Robotics and Intelligent Transportation Systems (RITS) Team, INRIA Paris\\ \and 
Spleenlab, Germany\\ \and
Valeo Vision Systems, Ireland
}
\maketitle              
\begin{abstract}
Lidar has become an essential sensor for autonomous driving as it provides reliable depth estimation. Lidar is also the primary sensor used in building 3D maps which can be used even in the case of low-cost systems which do not use Lidar. Computation on Lidar point clouds is intensive as it requires processing of millions of points per second. Additionally there are many subsequent tasks such as clustering, detection, tracking and classification which makes real-time execution challenging. In this paper, we discuss real-time dynamic object detection algorithms which leverages previously mapped Lidar point clouds to reduce processing. The prior 3D maps provide a static background model and we formulate dynamic object detection as a background subtraction problem. Computation and modeling challenges in the mapping and online execution pipeline are described. We propose a rejection cascade architecture to subtract road regions and other 3D regions separately. We implemented an initial version of our proposed algorithm and evaluated the accuracy on CARLA simulator.
\end{abstract}
\section{Introduction}
\label{sec:Introduction}
Autonomous driving systems are seldom complete nowadays without a controller, motion planning and perception stack. Pre-curated maps of their environments can be further used to improve the robustness and completeness. A modern mapping system, often referred as HD Vector Maps or Prior Maps, is typically comprised of the following stages:

\begin{itemize}
\item \textbf{Static Map extraction}: This is an offline step called mapping where a geometric 3D model is built for the static environment.
\item \textbf{Localization}: The vehicle localizes itself independently with a GPS and measures it's movement with an IMU. Alternatively the vehicle can localize itself  by map aligning within the known 3D map. 
\item \textbf{Dynamic events representation}: Besides providing a map of the environment, modern mapping systems also provide contextual information such as: speed limits, drivable directions and space, lane markings and distance to intersections. Dynamic critical events like accidents, roadworks and lane closures are additionally provided. Some systems are able to analyze curated driver behavior profiles to predict its dynamic behavior.
\end{itemize}

The current day mapping technologies are advancing rapidly, however a formal 
description of the pipeline/sequence of algorithmic operations is still lacking. 
Companies such as Civil Maps, HERE, DeepMap, TomTom and Mobileye have various 
implementations of 3D maps of the environment augmented with various temporal and 
spatial meta-data. Moreover, crowd-sourcing to learn the environment
\cite{dabeer2017end} has become an important trend in modeling and 
updating/maintaining these maps.

We briefly describe the outline of the paper. Section 1 describes why prior
3D maps are required and their essential characteristics. Section 2 Reviews
the current state of the art of Static 3D Maps and HD Maps, and the basic
point cloud representations. Section 3 describes the 3D mapping pipeline
while reviewing the literature regarding the various constituent steps.
Section 4 describes our experiments with the CARLA simulator : creation of
a evaluation framework for dynamic object extraction, creating a semantic
point cloud, and finally a visual demonstration of a bounding box based
point cloud inlier rejection. We conclude the paper with future work and
other operations essential to the completeness of 3D Maps.

\subsection{Motivation}
\label{subsec:Motivation}

Vehicle perception and interpretation of an unknown environment is a difficult and
computationally demanding task, where any prior information that contributes to augment
the information retrieved by the vehicle would aid to improve the performance.
As automated driving vehicles might be commissioned to drive over closed loop routes
and repeat pre-recorded paths, having a consistent parametric representation of the
static environment is very useful. This is particularly the case with automated
driver-less shuttles or taxi services. Current experimental driver-less prototypes
commonly use globally registered 3D prior-maps with centimetric-precision. They are created using Lidar sensors and GPS+INS systems \cite{levinson2011towards}.

Point clouds retrieved by Lidar provide an accurate geometrical representation of the
environment. However, they do not explicitly provide information about unknown areas or
free space. They are also memory-intensive and lack an inherent mechanism to adapt to
changes in the environment \cite{wurm2011hierarchies}.

For these reasons, a pre-recorded 3D representation of the static surroundings of the environment is a useful prior that can be used for localization, obstacle detection and tracking tasks. Such representation would ideally be characterized by the following features:

\noindent \textbf{Accurate 3D Model representation} of the static background environment, including driving road surfaces, buildings planes/facades, lamps, roundabouts, traffic signs among others. Here, obstacles can be classified into two categories based on their stationarity across the training and test set : 
\begin{enumerate}
    \item \textbf{Stationary Objects (SO)} :  Completely stationary scene components such as the road surface and buildings.
    \item \textbf{Non-Stationary Static Objects (NSSO)} : Static objects that appeared or disappeared between mapping and re-localization steps.
\end{enumerate}
An online background estimation and removal would help to reject a considerable portion of static points in the cloud to localize dynamic objects in the scene.

\noindent \textbf{Computation Speedup} : Lidar point clouds (64 Layers) usually contain around 100K points per frame, with approximately 10 frames/sec. Computation can be reduced by performing clustering of point clouds into simpler higher level models like planes. For example, plane models can be fit during mapping stage for road surfaces, building facades, etc and bounding boxes can be used for volumetric clusters estimated. This way, obstacle detection and tracking steps in the perception pipeline can be performed faster. The computational speedup is also dependent on the data structures adapted and used to store the data, we refer the readers to work done in \cite{yin2017mastering}.  

\noindent \textbf{Updatability}. Apart from the cost of 3D Lidar sensors, a significant cost of a successful automated driving system goes into building and maintaining an updated 3D prior map. Surveying environments on a large scale requires keeping account of changes in the road surface, buildings, parked vehicles, local vegetation and seasonal variation, which is prohibitively expensive for a map provider. Any economic benefits of selling automated driving systems may be offset by the constant need to re-survey the environment where they operate in.  Changing environments are handled in \cite{churchill2012practice,maddern2015leveraging}, where instead of building a single static global map, each vehicle independently creates and maintains a set of partial updates. The confidence of the parametric representation increases with the number of observations and updates done by each vehicle.

\section{Prior Maps and Point Cloud Representation} 

\subsection{Types of Maps}

Mapping is one of the key pillars of automated driving. The first reliable demonstrations  of automated driving by Google were primarily reliant on localization to pre-mapped areas.  Because of the scale of the problem, traditional mapping techniques are augmented by  semantic object detection for reliable disambiguation. In addition, localized high  definition maps (HD maps) can be used as a prior for object detection. More details of mapping and usage of modern deep learning methods are discussed in \cite{milz2018visual}. \\

\textbf{Private Small Scale Maps:} There are three primary reasons for the use of customized small scale maps. The first reason is privacy where it is not legally allowed to map the area, for example, private residential area. The second reason is that HD maps still do not cover most of the areas. The third reason is the detection of dynamic structures, that may differ from global measurements. This is typically obtained by classical semi-dense point cloud maps or landmark based maps. \\

\begin{figure}[!t]
\centering
\includegraphics[width=0.8\textwidth]{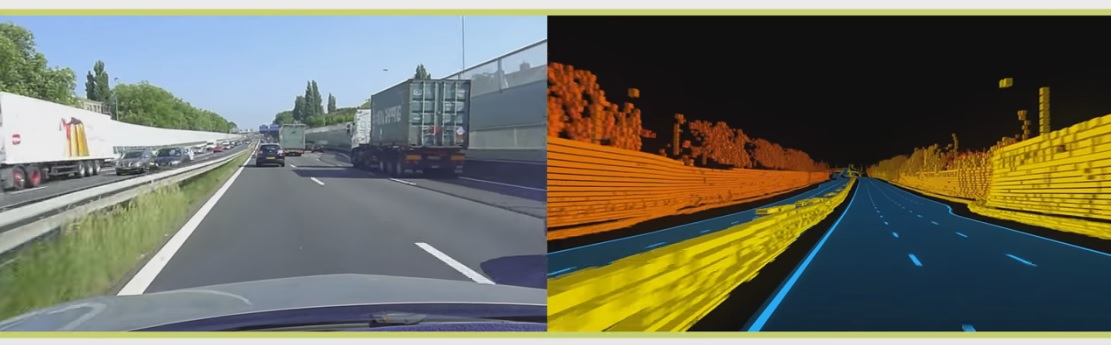}
\caption{Example of High Definition (HD) map from TomTom RoadDNA (Reproduced with permission of the copyright owner)}
\label{fig:HDmaps}
\end{figure}

\textbf{Large Scale HD Maps:} There are two types of HD maps namely Dense Semantic Point Cloud Maps and Semantic Landmark based Maps. Semantic Landmarked based maps are an intermediate solution to dense semantic point cloud and likely to become redundant.

\begin{enumerate}
    \item Landmark based Maps are based on semantic objects instead of generic 3D point clouds. Thus it works primarily for camera data. Companies like Mobileye and HERE follow this approach. In this method, object detection is leveraged to provide an HD map and the accuracy is improved by aggregating over several observations from different cars.
    \item Dense Semantic Point Cloud Maps: This version is the best representation where all the semantics and dense point cloud are available at high accuracy. Google and TomTom follow this approach. Figure \ref{fig:HDmaps} demonstrates dense semantic 3D point cloud from TomTom and alignment to an image. However this approach is computationally expensive and needs large memory requirements. In this case, mapping is treated as a stronger cue than perception. If there is good alignment with the map, all the static objects (road, lanes, curb, traffic signs) are obtained from the map already and dynamic objects are obtained via  background subtraction. In this work, we propose an efficient solution to obtain dynamic objects.
\end{enumerate} 


\subsection{Review On Grid-based Representations}

Grid-based representations split the space into equally sized cells in 
order to represent the state of specific portions of the environment. Moravec et al.
\cite{Moravec-1985-15232}, create a 2D grid based occupancy map with a sonar. Herbert et al. 
\cite{Herbert100111} proposed to extend this representation by assigning an additional 
variable to store at each cell the height of objects above the ground level, this approach is 
usually referred as an elevation map. Moreover, a multiple surface representation was 
presented by \cite{Triebel4058725}, where several height values can be stored within the 
cells.


The main advantage of grid-based representations is that free, occupied and unknown space can be represented from the measurements obtained by a range sensor, usually a Lidar. These methods are commonly known as occupancy-grids, where a ray casting operation \cite{Bresenham5388473} is performed along each measurement in order to decrease the occupancy probability of traversed cells and increase it for the the impacted ones \cite{Thrun:2005:PR:1121596}. This is however a costly operation. Each cell is recursively updated by applying a static state binary Bayes filter as introduced in \cite{Elfes30720}, where the occupancy probability of a cell $v$ being occupied $p(v|\mathbf{z}_{t,1:n})$ is defined as:  

\begin{equation}\label{eq:OccupancyGridUpdate}
p(v|\mathbf{z}_{t,1:n}) = \bigg[1+ \frac{1-p(v|\mathbf{z}_{t,n})}{p(v|\mathbf{z}_{t,n})}\frac{1-p(v|\mathbf{z}_{t,1:n-1})}{p(v|\mathbf{z}_{t,1:n-1})}\beta \bigg]^{-1} , \ \ \beta = \frac{p(v)}{1-p(v)}
\end{equation}


\noindent where $\mathbf{z}_{t,1:n}$ is the complete set of 
sensor measurements $\{\mathbf{z}_{t,1}, ..., \mathbf{z}_{t,n}\}$ 
obtained from the sensor returns at time $t$.The term $\beta$ 
depends on the prior knowledge about the state of the cell $v$, 
as shown in Eq. (\ref{eq:OccupancyGridUpdate}). If this initial state 
is unknown, then $p(v) = 0.5$ and $\beta = 1$. The updates 
are then performed by following:

\begin{equation}\label{eq:OccupancyState}
p(v|\mathbf{z}_{t,i}) =
\begin{cases}
p_{free}\text{,} & \text{the ray $\mathbf{z}_{t,i}$ traverses through $v$} \\
p_{occupied}\text{,} & \text{the ray $\mathbf{z}_{t,i}$ impacts within $v$} \\
0\text{,} & \text{otherwise}
\end{cases}
\end{equation}

\noindent where $p(v|\mathbf{z}_{t,i})$ corresponds to the probability update given by observation $\mathbf{z}_{t,i}$ and $p_{free}$ and $p_{occupied}$ are the assigned probabilities when a cell is completely traversed or impacted respectively. A wide range of approaches \cite{ushani2017learning,Doherty2017BayesianGK,Wurm10octomap} chose to perform this update using a log-odds ratio formulation $L$, where instead of multiplying terms from prior, past and current measurements as shown in Eq. (\ref{eq:OccupancyGridUpdate}), logarithmic properties allow to perform this update with simple additions:

\begin{equation}
\label{eq:LogOccupancyGridUpdate}
L(v|\mathbf{z}_{t,1:n}) = \left( \, \sum_{i=1}^{n} L(v|\mathbf{z}_{t,i}) \, \right) - L(v) ,
\ \text{where} \ L(a) = \log\left(\frac{p(a)}{p(\neg a)}\right)   
\end{equation}

Similarly as in Eq. (\ref{eq:OccupancyState}), log-odds probability values $l_{free}$ and $l_{occupied}$ are defined for the update of $L(v|\mathbf{z}_{t,i})$. The advantage of this approach is that truncation issues associated with probabilities close to $0$ or $1$ can be avoided.

\section{3D-Prior Maps Pipeline for Online Rejection }

\begin{figure}[t!]%
\centering
\scalebox{.8}{%
\begin{tikzpicture}
\definecolor{ochre}{RGB}{191,144,0}
\draw[rounded corners, dashed, thick, color=ochre, fill=ochre!10] (1.5,0) rectangle (10.3, 2.5) {};
\draw (0,1) node (a) {$X=\{x_i\}_{i=1}^N$};
\draw[->, ultra thick] (1.2,1) -- (1.8,1);
\draw[thick] (1.8,0.3) rectangle (4.3,1.5) node[pos=.5,align=center]{Ground plane\\extraction};
\draw[->, ultra thick] (4.3,1) -- (4.9,1);
\draw[thick] (4.9,0.3) rectangle (7.1,1.5) node[pos=.5,align=center]{Volumetric\\clustering};
\draw[->, ultra thick] (7.1,1) -- (7.7,1);
\draw[thick] (7.7,0.3) rectangle (10,1.5) node[pos=.5,align=center]{Main planes\\estimation};
\draw[->, ultra thick] (10,1) -- (10.6,1);
\draw (11.8,1) node (b) {$\Theta=\{\theta_j\}_{j=1}^K$};
\draw (5.9,2) node (c) {\textbf{Mapping stage}};
\draw (0,0.2) node[align=center] (d) {\shortstack{{\scriptsize~Set of}\\{\scriptsize~point clouds}}};
\draw (11.9,0.2) node[align=center] (d) {\shortstack{{\scriptsize~Set of main planes'}\\{\scriptsize~parameters}}};
\end{tikzpicture}%
}\\%
\vspace{7pt}%
\scalebox{.7}{%
\begin{tikzpicture}
\definecolor{green}{RGB}{84,130,53}
\draw[rounded corners, dashed, thick, color=green, fill=green!25] (1.5,0) rectangle (12.8, 4.8) {};
\draw (0,3) node (a) {$X=\{x_i\}_{i=1}^N$};
\draw[->, ultra thick] (1.2,3) -- (1.8,3);
\draw[thick] (1.8,2.3) rectangle (4.9,3.7) node[pos=.5,align=center]{\textit{Planar}\\rejection cascade\\classifier};
\draw[->, ultra thick] (4.9,3) -- (5.5,3);
\draw[thick] (5.5,2.3) rectangle (8.6,3.7) node[pos=.5,align=center]{\textit{Volumetric}\\rejection cascade\\classifier};
\draw[->, ultra thick] (8.6,3) -- (9.2,3);
\draw[thick] (9.2,2.3) rectangle (12.5,3.7) node[pos=.5,align=center]{Non-stationary\\static/dynamic\\obstacle classifier};
\draw[->, ultra thick] (12.5,3) -- (13.1,3);
\draw (14.3,3) node (b) {$V_{NSSO}, \ V_{DO}$};
\draw (7,4.3) node (c) {\textbf{Driving stage}};
\draw (0,2.3) node[align=center] (d) {\shortstack{{\scriptsize~Set of}\\{\scriptsize~point clouds}}};
\draw (14.3,2) node[align=center] (e) {\shortstack{{\scriptsize~Non-stationary}\\{\scriptsize~static and dynamic}\\{\scriptsize~objects classified}}};
\draw[->, thick] (3.3,1.7) -- (3.3,2.3);
\draw (3.3,1.1) node[align=center] (f) {\shortstack{$\Theta$\\{\scriptsize~Set of main}\\{\scriptsize~planes' parameters}}};
\draw[->, thick] (7,1.7) -- (7,2.3);
\draw (7,0.9) node[align=center] (g) {\shortstack{$\Theta$\\{\scriptsize~Set of}\\{\scriptsize~volumetric}\\{\scriptsize~cluster parameters}}};
\draw[->, thick] (5.2,3) -- (5.2,2);
\draw (5.2,1.4) node[align=center] (g) {\shortstack{$O$\\{\scriptsize~Obstacle}\\{\scriptsize~set}}};
\end{tikzpicture}%
}

\caption{A Standard clustering, Ground Plane extraction, obstacle detection 
and tracking pipeline in an autonomous driving pipeline.}
\label{fig:pipeline}
\end{figure}
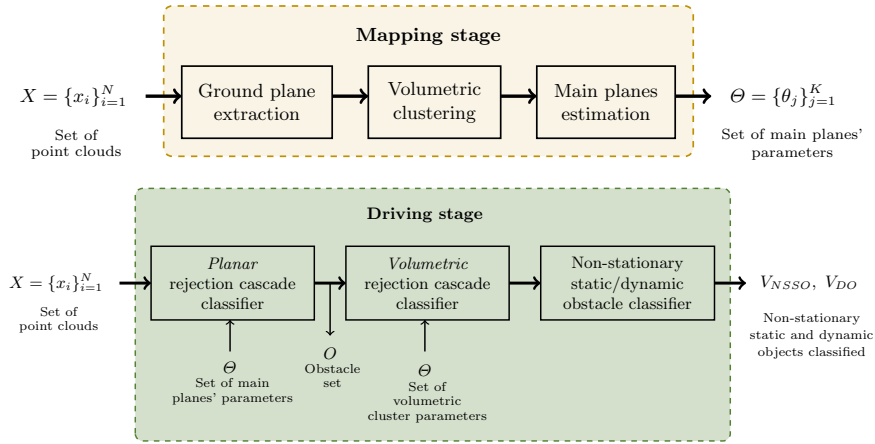

3D-Prior Maps built in various pipelines essentially contain two steps: Initially, an
offline \textbf{Mapping} stage is performed, during which the computationally heavy 
parameter estimation is carried out. An online \textbf{Driving} stage constitutes the prediction of planar and volumetric inliers w.r.t the mapping phase estimated parameters. Figure \ref{fig:pipeline} demonstrates a concrete example of parameter estimation during the offline mapping stage and online model inlier rejection during the online driving stage. Similar pipelines are developed in \cite{asvadi20163d} and \cite{asvadi2016two}. In \cite{sixta2017lidar} a voxelization step is included to reduce the time complexity for the clustering/detection in the down-streams stages.  Further on, we refer to the parameter estimation stage as training, and the inlier rejection/model fitting stage as testing. The point cloud dataset acquired for training is assumed to have no dynamic obstacles, while the test dataset shall contain point clouds acquired at new vehicle poses (thus incurring errors in GPS localization and vehicle orientation) with a set of dynamic objects observable in each frame.

The obstacle detection pipeline can be then considered as a binary classification problem, with the background being constituted by the ground and building facades planes as well as volumetric static obstacles (vegetation, parked cars among others), while the foreground is formed by the dynamic obstacles not belonging to the background. Since the parametric estimation for geometric and volumetric obstacles is performed on a training set containing no moving obstacles, while prediction is performed on unseen test set, one could detect non-stationary static obstacles (NSSO) in the test set which have appeared/disappeared in between the mapping and driving steps.

We categorize the 3D object mapping problem in two categories : 
First is to consider the detection of 3D obstacles on prior maps as a stationary background estimation problem, which is well known in the background subtraction community. The second is to separate the background objects in geometrical and volumetric structures such as planes, road surfaces and dense clusters. We shall explore the second approach in our paper. The key steps performed in this approach are:

\begin{enumerate}
\item \textbf{Global Coordinate Point Cloud Registration}: Each point cloud frame produced by the vehicles across different poses is registered into a common global origin, yielding a single point cloud representation of the entire environment. This is usually achieved by applying rigid transformations to each point cloud frame with the aid of GPS/INS and Inertial Measurement Systems (IMU). Registration algorithms are also applied for this purpose, where Iterative Closest Point (ICP) algorithm is the most popular approach \cite{bellekens2014survey}. 

\item \textbf{Ground Surface Approximation}: Surface reconstruction of roads and normal estimation to obtain their local orientation. Planar ground assumptions are usually considered to estimate a planar surface parameters. Non planar surfaces can be approximated with several planes as in \cite{asvadi20163d}.

\item \textbf{Planar surface extraction}: Extract other dominant planes present in the geometric representation, commonly linked to building facades.

\item \textbf{Clustering and Classification}: Clustering of the remaining volumetric features (vegetation, parked vehicles, lamp posts, etc.) in the global map coordinate frame. The classification step is usually performed over the clusters to obtain a class for each cluster : Vehicles, Buildings, Vegetation and others. This provides semantic information regarding the type of obstacle.

\item \textbf{Long-term occupancy probability estimation}: In this step the environment 
is classified in the mentioned classes: Stationary Objects (SO), Non-Stationary Static 
Objects (NSSO) and dynamic obstacles. The precision of the classification will depend on 
the amount of gathered data. NSSO clusters are removed or updated based on user 
intervention. For example, a construction site might have appeared during the mapping
stage, while disappeared during the online driving stage. In such a case, the pipeline
classifies the empty construction site as a dynamic obstacle.

\end{enumerate}

\subsection{Ground surface approximation}
\label{subsec:GroundSurfaceApproximation}

Road surface extraction allows the vehicle to detect the drivable space while removing a considerable proportion of points for the subsequent obstacle detection and classification steps. Random sampling methods (RANSAC), normals estimation or Hough based methods are among the most popular approaches for this task.

Recently, Chen et al. \cite{chen2017lidarhistogram} build a depth image by
using the Lidar spherical coordinates, where each row (i.e., each fixed azimuth
value $\phi$) corresponds to a given laser layer of the sensor. The columns of
the image correspond to the horizontal polar angles ($\theta$) and the image
intensities are represented by the radial distances ($d$). An example of the
depth image can be seen at Fig. \ref{fig:lidar_hist_ex1} top left. Following the 
assumption that for a given sensor layer (a given $\phi$) all the points in the 
road will be at a same distance from the sensor along the $x$ axis, the
$v$-disparity of a given pixel $(p,q)$ is calculated as:

\begin{equation}
\label{eq:LidarHistogram}
\Delta_{p,q} = \frac{1}{x(p,q)}=\frac{1}{d_{p,q}\cos(\phi_{p,q})\cos(\theta_{p,q})}
\end{equation}

Since all road points in a given row $q$ will share a similar value $x$, they will 
fall into a same box in the histogram and, assuming the road is flat, the 
high-intensity values in the histogram will draw a line (since the value of 
$\Delta_{p,q}$ will change linearly as we traverse the values of $\varphi$). 
Aside from road plane parameter estimation, this method can also be used for 
online obstacle detection, due to the lightness of its computations. All objects 
placed on the road will be closer than the road to the sensor, and will therefore
have a higher disparity value. Thus, they will be located above the line that
corresponds to the road in the histogram (see fig. \ref{fig:lidar_hist_ex1}).

\subsection{Static Background Estimation in 3D-Environments}

\noindent \textbf{Plane extraction : RANSAC based methods}
A standard plane extraction algorithm is the Random sample consensus (RANSAC)
method. This algorithm aims to find a plane in the scene, where the resulting
plane should "contain" the most possible points with respect to a threshold
distance from plane to point. In the self-driving context, the road is therefore
assimilated to a plane and we use a large threshold to accommodate for any
potential curvature. Fast Ground Segmentation extraction methods on Lidar Data
with the use of Squeeze-net Architectures are capable of real-time prediction 
performances \cite{velas2018cnn}.

\noindent \textbf{Geometric approximation}:
Plane extraction methods can be categorized into Hough-based, region growing, 
or RANSAC-based approaches. Authors \cite{grant2013finding} describe a novel 
Hough-based voting scheme for finding planes in 3D point clouds. They assume an 
important geometrical property of the Lidar, i.e. planes in the 3D intersect 
with Lidar rays at a fixed azimuth angle $\phi$ in conic sections of varying 
curvature dependent upon the inclination angle of the Lidar and the relative 
orientation of the plane. A fast approximation of regions where for a 1D 
scan $d_\phi(\theta)$ smoothly varying segments are extracted, followed by 
accumulator framework similar to the one in randomized Hough transform is 
utilized to vote the planes that were extracted. Background Subtraction based 
approaches have been studied in accessibility analysis using Evidence Grids
\cite{anderson2011background}.

\subsection{Review on point cloud representations for Mapping}
Here we shortly review the different classes of methods to evaluate
a static 3D Map. Readers are directed towards a recent comprehensive 
review on road object extraction using laser point cloud data by authors
\cite{reviewRoadObjectsPointclouds2018}.
\noindent \textbf{Multi-resolution and multi-scale methods}:
In \cite{wolcott2016robust} authors model each cell of an occupancy grid 
with a multi-resolution Gaussian Mixture models (GMMs) for obstacle mapping 
as well as localization in prior maps. In \cite{einhorn2011finding}, a 
dynamic method to adequate the resolution of the grid mapping-cell sizes 
by using an octree structure is presented. This could be applied on the 
prior maps to help compress the amount of data and to improve the obstacles 
detection performance.

\noindent \textbf{Compressing Information} : Authors in
\cite{nelson2015environment} provide a principled study on how 
to compress map information for autonomous exploration robots. To reduce
the computational cost of exploration, they develop an information
theoretic strategy for simplifying a robot’s representation of the
environment, in turn allowing information-based reward to be evaluated more
efficiently. To remain effective for exploration, their strategy must adapt
the environment model in a way that sacrifices a minimal amount of
information about expected future sensor measurements.

\noindent \textbf{Dynamic Obstacle Classification methods}:
We shortly describe methods that do no rely on estimating an accurate 3D 
obstacle prior map, but instead extract features that classify the scene 
into foreground and background. Authors in \cite{ushani2017learning} 
create a logistic classifier is built on a binary feature map created 
from the voxelized point cloud grid. The training of the binary 
classifier is performed on the KITTI dataset with positive class 
for dynamic obstacles given by the bounding box tracklets, while 
the background is any window not containing a dynamic obstacle. 
\cite{simon2018complex} use the region proposal network from YOLO
to perform bounding box location and orientation prediction.
MODnet \cite{siam2017modnet} learns to directly extract moving objects 
using motion and appearence features and could be used to augment 
existing dynamic object extraction methods. 

\noindent \textbf{Multi-model approaches}:
In \cite{ortega2011segmentation}, multi-modal background subtraction is performed
by classifying image data as background/foreground. This is done by employing a
per-pixel Gaussian mixture model estimated on a background without 
dynamic obstacles. The lidar point clouds are reprojected into the camera's
field of view and consequently masked as foreground or background by 
applying the pre-existing correspondence between pixels and lidar-points.
This methodology provides a direct bridge to a family of background 
subtraction methods \cite{bouwmans2014traditional} that could now be 
applied on the image data to perform binary classification
of lidar points.

Authors in \cite{wolcott2014visual} utilize a lidar based mapping 
stage that not only use the Lidar point clouds coordinates $x, y, z$, 
but also the reflectance $\alpha$. They localize a monocular camera 
during the driving stage within a 3D prior map by evaluating the 
normalized mutual information between the illuminance with surface
reflectivities obtained during the mapping stage, and localize 
themselves given an initial pose belief.

\noindent \textbf{Deep Learning based methods}
Authors in \cite{wang2018dels} propose a deep learning based system for fusing
multiple sensors, i.e. RGB images, customer-grad GPS/IMU, and 3D semantic maps, 
which improves the robustness and accuracy for camera localization and
scene parsing. Authors also propose the Zpark dataset along with a study 
that includes dense 3D semantically labeled point clouds, ground truth camera 
poses and pixel-level semantic labels of video camera images,
which will be released in order to benefit related researches.

Terrestrial point clouds provide a larger field of view as well as point cloud density.
3D point cloud semantic segmentation produced with terrestrial scanners have 
been evaluated by authors in \cite{hackel2017isprs}. The 3D prior maps mapping 
architecture could also be augmented by aligning ground scanned point clouds 
with Airborne point clouds. Authors in \cite{elbaz20173d} register aerially scanned 
point cloud and local ground scanned point clouds within Airborne point clouds 
using Deep stacked Autoencoders. Authors in \cite{sharma2016vconv} propose a 
full convolutional volumetric autoencoder that learns volumetric representation 
from noisy data by estimating the voxel occupancy grids. 

\begin{figure}
\includegraphics[width=0.45\linewidth]{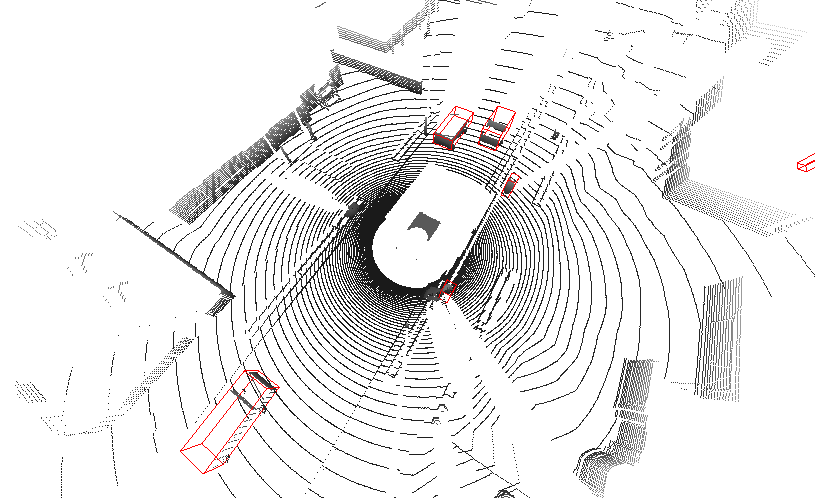}
\includegraphics[width=0.45\linewidth]{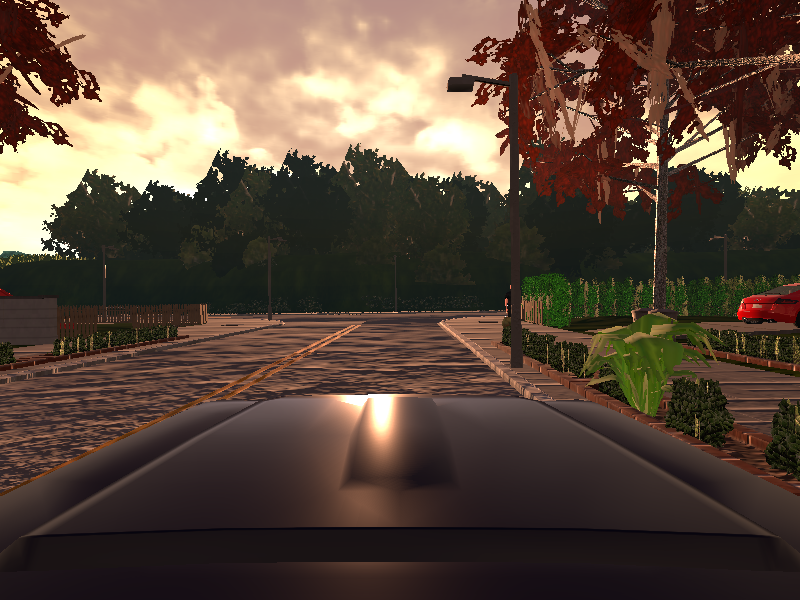}
\caption{Figure (left) shows tracklets extracted from CARLA simulator. Figure(right) 
shows the corresponding view of camera.}
\label{fig:tracklets}
\end{figure}

\section{Experiments}
We use the CARLA simulated cities with dynamic objects as our test framework.
During the \textbf{Mapping stage}, point clouds at each vehicle pose (GPS coordinate 
with vehicle orientations yaw, pitch and roll) are obtained from the CARLA simulator.
In the driving stage, we perform three distinct steps : Firstly, we extract a parametric
representation of the road, namely the RANSAC and the Lidar Histogram as described before. 
Secondly, we cluster the current frame with the hierarchical DBSCAN clustering algorithm 
\cite{McInnes2017,Campello2015HDBSCAN}. The third step involves the estimation of 
the final rejection cascade model using robust bounding boxes for planar clusters, 
while minimum volume bounding boxes for volumetric clusters. 

\subsection{CARLA Setup}
The CARLA simulator \cite{Dosovitskiy17} has recently created a lidar point cloud 
generator that enables users to parametrise the simulated sensor with different 
parameters such as its angular resolution, number of layers among others. Virtual 
KITTI based point clouds are now made available by authors in 
\cite{3dsemseg_ICCVW17}. On the other hand, \cite{yue2018lidar} now provide simulated 
Lidar point clouds and object labels, aiming to augment the datasets to be used for 
supervised learning using deep convolutional networks.

In our study, we use the CARLA simulated lidar point cloud stream, with the
localization and IMU information generated for the simulated vehicles to map
the environment. CARLA provides a reliable environment where the performance 
of the dynamic objects detection can be tested, since the user has
access to the location, speed and bounding boxes of all agents in the scene.
Figure \ref{fig:tracklets} shows an example of the object tracklets with the 
associated camera image.The semantic segmentation of the scene is also 
provided.  

\begin{figure}
    \centering
    \includegraphics[width=0.95\linewidth]{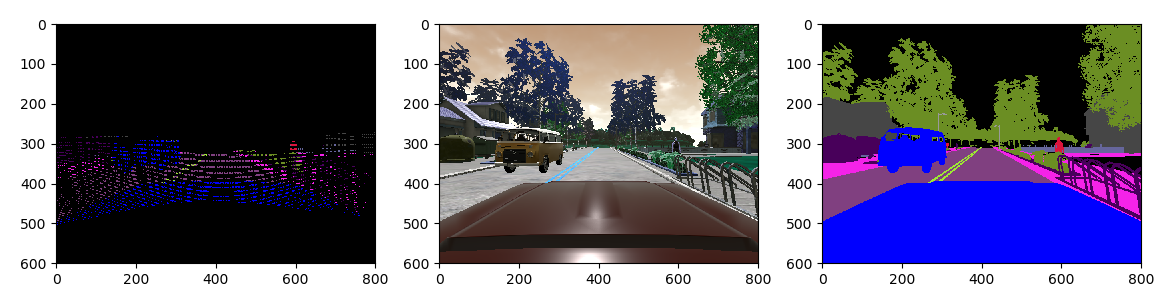}\\
    \caption{Left to right : Pointcloud with labels in camera view, image from
    camera, semantic segmentation of the scene.
    Mapping the semantic segmentation labels to 3D point cloud from 
    the Lidar. These initial results point towards the possibility of obtaining 
    a completely labeled semantic point cloud for downstream classification tasks 
    and parameter cross-validation purposes.}
    \label{fig:lidar2semanticseg}
\end{figure}

\subsection{Clustering and Road Extraction}
The road extraction in CARLA is performed using a frame based RANSAC plane estimation. 
The Lidar Histogram has been observed to be useful in cases where the road surface is 
not a plane and contains positive and negative slopes.

\begin{figure}
	\centering
    \includegraphics[width=0.45\textwidth]{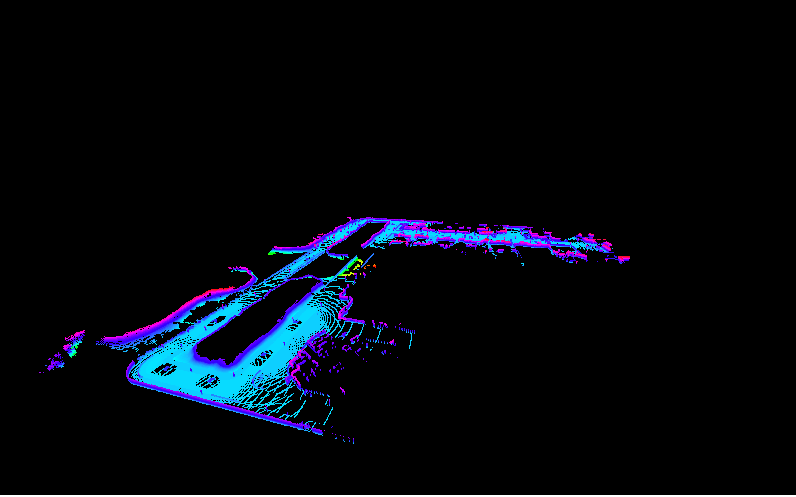}
    \includegraphics[width=0.45\textwidth]{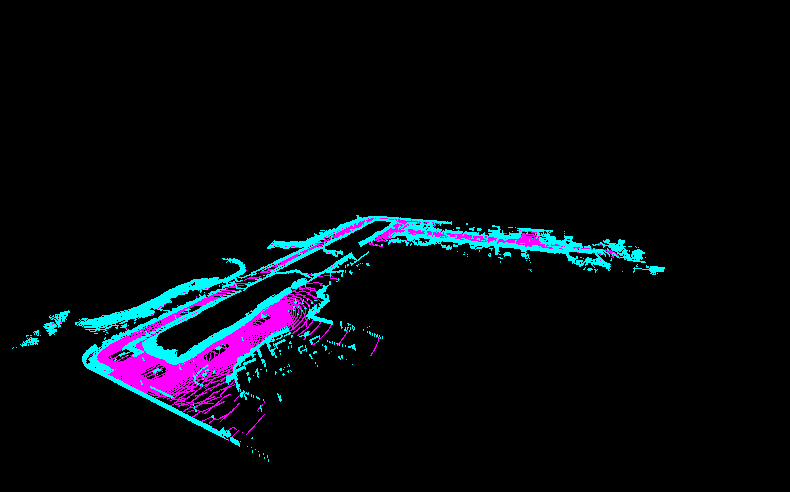} \\
    Aligned point cloud \hspace{2.5cm} Road Extraction with RANSAC\\
    \includegraphics[width=0.45\textwidth]{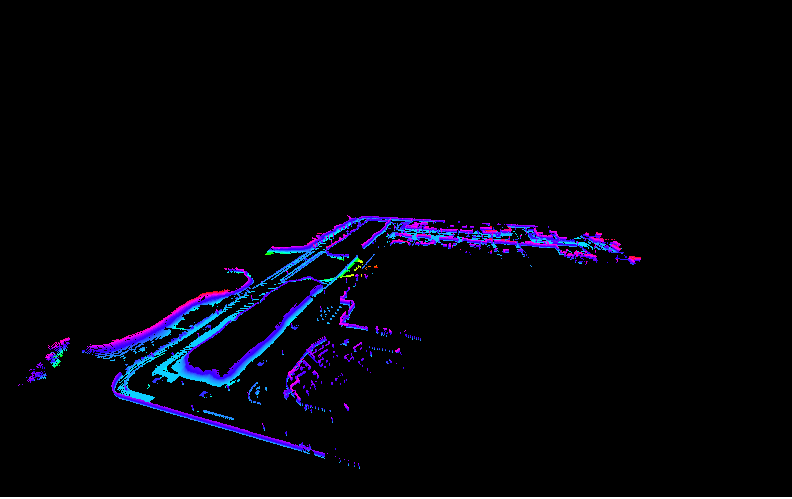}
    \includegraphics[width=0.45\textwidth]{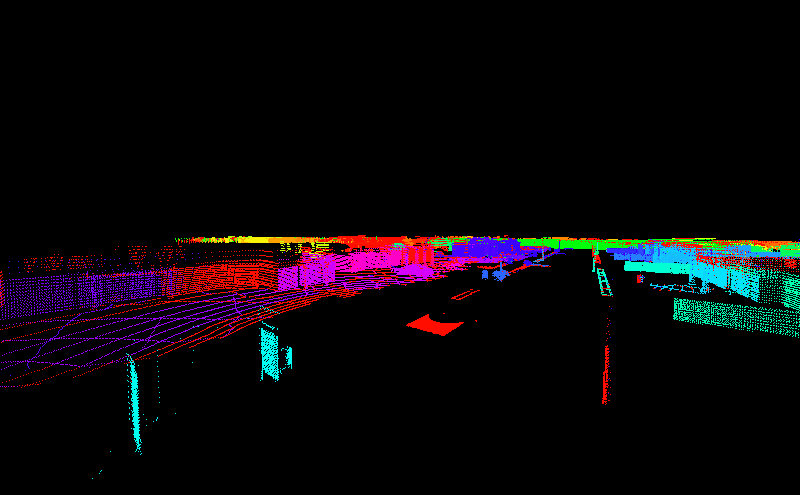}\\
    Point cloud without Road \hspace{2cm} Globally Clustered pointcloud\\
    \caption{Joined cloud in global coordinates, Road extraction using
    RANSAC, Road removed point cloud, Clustered point cloud using the
    hierarchical DBSCAN (HDBSCAN) on each frame followed super-clustering
    step on centroids from the previous step.
    \cite{McInnes2017}.}
    \label{fig:full_frames}
\end{figure}

The clustering process involves two steps : An initial frame-based clustering is achieved 
using the HDBSCAN algorithm \cite{Campello2015HDBSCAN}, as show in figure \ref{fig:full_frames}, 
and a subsequent clustering of the centroids of each frames cluster, to obtain a globally 
consistent cluster label. This second clustering step, termed as super-clustering is 
performed again using the HDBSCAN. Our primary motivation in using HDBSCAN is to 
handle the variable density of point-clouds due to the geometrical projection produced by the 
Lidar sensor.

After each step of the frame based clustering a the labelled point cloud is saved back for each frame.
Once all frames have been labelled, a global association step is performed in the final step when 
the centroids across all frames are clustered to obtain a consistently labelled point cloud. In the future, a hierarchical data structure could alleviate this two-level clustering task.

\subsection{Alternative road extraction with Lidar-histogram}

\begin{figure}[h]
    \centering
    \includegraphics[width=0.9\textwidth]{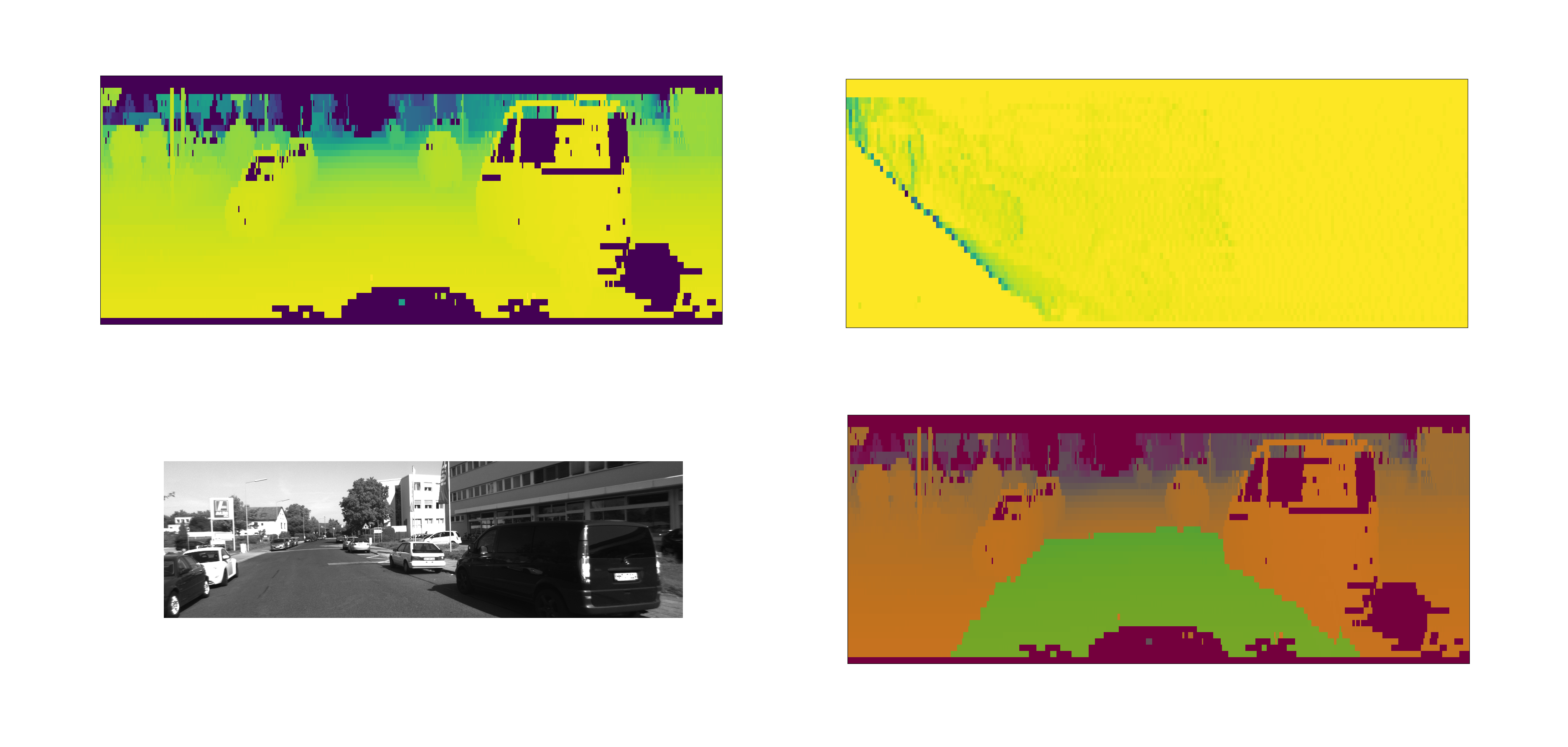}
    \caption{Example of the LIDAR-histogram method at work, applied to a frame from the KITTI dataset (drive 9, frame number 228).  The pictures above show the depth image (left) and histogram (right). The images below are the original picture and the result of the the classification of the pixels (the ones corresponding to the road in green).}
    \label{fig:lidar_hist_ex1}
\end{figure}

In this section we show the result of road segmentation on readings from 
the KITTI dataset, results are shown in figures \ref{fig:lidar_hist_ex1}.
CARLA simulated roads in the default environments are flat, white KITTI 
provides roads with changes in gradient.

Since the Velodyne points in this dataset are given in Cartesian
coordinates, they were transformed to spherical ones first. We used a naive
approach to approximate the line representing the road, selecting a desired
number of the most intense values and fitting a line with the least squared
error. Although this proved to be precise enough, the line could also be
fitted using a RANSAC-based method or even Hough transforms. In order to
get the road points from the line, both the disparity and $x$ values were
computed for each row and all values within a threshold were accepted.
The pixels corresponding to the road have correctly been identified, as can
be seen in both figures, without overriding the obstacles present in the
lane. If desired, the classified point clouds in Cartesian coordinates
could then be retrieved by putting the points in spherical coordinates back
to Cartesian ones.

\subsection{Planar and Volumetric cluster classification}
\begin{figure}
    \centering
    \includegraphics[width=0.45\textwidth]{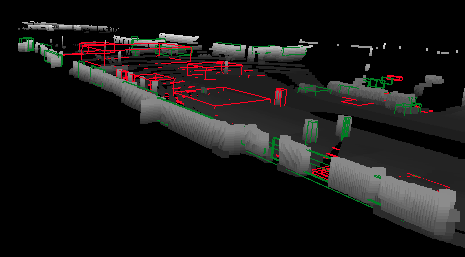}
    \includegraphics[width=0.45\textwidth]{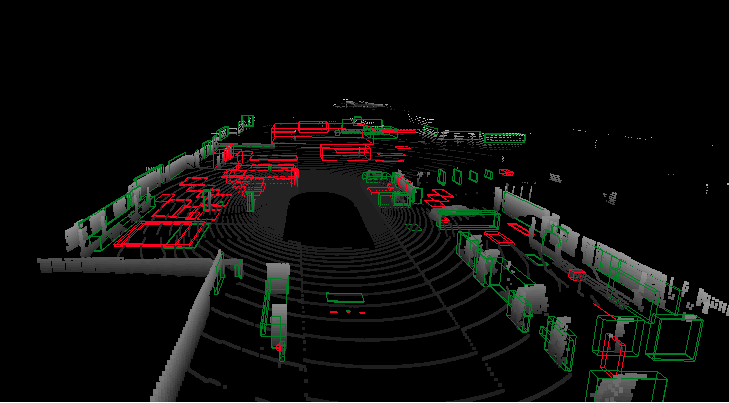}
    \caption{A prior 3D map constituting of two classes of bounding boxes : Robust 
    planar bounding boxes for planes (green), and volumetric minimum volume bounding box (red).
    In future studies we plan to use these bounding boxes to achieve a rejection cascade.}
    \label{fig:3dmap}
\end{figure}

In our third step after road extraction using RANSAC and frame based 
clustering using HDBSCAN, we describe briefly how we obtain a inlier
rejection bounding box for such clusters. The planar and volumetric
clusters are decided by a threshold on average planarity feature. The
planarity is obtained from eigenvalues of the 3D structure tensor can given
by $P_\lambda = (\lambda_2 - \lambda_3)/\lambda_1 \in [0, 1]$, for each
point in the point cloud. We show two types of clusters with their
respective bounding boxes in figure \ref{fig:3dmap}.

\section{Conclusions}
3D-Prior maps for obstacle detection has become a key engineering problem in 
today's autonomous driving system, moving a large part of the detection problem 
into the construction of precise 3D probabilistic descriptions of the environment. 
In this study, we reviewed the basic steps involved in the construction 
of prior maps : Road extraction, clustering and subsequent plane extraction,
super-clustering for globally consistent clusters.

We demonstrated the results of applying Lidar Histogram to various tracks in 
the KITTI dataset. We showed it to be an efficient parametric representation 
of the road surface, while modeling positive and negative (holes) obstacles. 
It also provides the decision making system with the estimation of 
free space around dynamic obstacles. Decomposing the road segmentation algorithm 
into off-line mapping and online driving stages is key to obtain real-time 
performance while not conceding accuracy.

Estimating the frequency of occurrence of different geometrical and 
volumetric features enables us to envisage an efficient implementation 
following a rejection cascade employed first by \cite{viola2001rapid}. 
Our implementation follows background subtraction based cascade as 
developed in \cite{kiran2017real}.

Furthermore maintaining the temporal relevance of obstacles in 3D-Maps 
is a costly and an important question. Long-term mapping solutions that
assume a semi-static environment are now in development
\cite{rosen2016towards}. Real-time implementations should include a 
third phase aside Mapping and Driving, that is an efficient change-point
detection (in the point clouds associated with non-stationary static
obstacles) and parameter update mechanism.

%
%
%
%
\bibliographystyle{splncs04}
\bibliography{bibliography}

\end{document}